\documentclass[10pt,twocolumn,letterpaper]{article}

\usepackage[pagenumbers]{cvpr} 

\definecolor{cvprblue}{rgb}{0.21,0.49,0.74}
\usepackage[pagebackref,breaklinks,colorlinks,allcolors=cvprblue]{hyperref}

\newcommand{\method}{\texttt{HBridge}\xspace}

\definecolor{Gray}{gray}{0.94}
\definecolor{liGray}{gray}{0.5}
\definecolor{LightCyan}{rgb}{0.88,1,1}

\usepackage{colortbl} 
\usepackage{xcolor}

\usepackage{wrapfig}
\usepackage[capitalize]{cleveref}  
\crefname{section}{Sec.}{Secs.}
\Crefname{section}{Section}{Sections}
\crefname{table}{Tab.}{Tabs.}
\Crefname{table}{Table}{Tables}
\crefname{figure}{Fig.}{Figs.}
\Crefname{figure}{Figure}{Figures}
\crefname{equation}{Eq.}{Eqs.}
\Crefname{equation}{Equation}{Equations}
\newlength\savewidth\newcommand\shline{\noalign{\global\savewidth\arrayrulewidth
  \global\arrayrulewidth 1pt}\hline\noalign{\global\arrayrulewidth\savewidth}}

\definecolor{myblue}{RGB}{224, 224, 224}

\def\paperID{2829} 
\def\confName{CVPR}
\def\confYear{2026}

\title{
HBridge: H-Shape Bridging of Heterogeneous Experts for\\Unified Multimodal Understanding and Generation
}

\author{
   \hspace{-0.4cm} 
   Xiang Wang$^{1}$
     \hspace{0.01cm} 
    Zhifei Zhang$^{2}$
     \hspace{0.01cm} 
     He Zhang$^{2}$
     \hspace{0.01cm} 
     Zhe Lin$^{2}$
     \hspace{0.01cm} 
     Yuqian Zhou$^{2}$
     \hspace{0.01cm} 
     Qing Liu$^{2}$
     \hspace{0.01cm} 
     Shiwei Zhang$^{1}$
     \hspace{0.01cm} 
     Yijun Li$^{2}$ \\
     \hspace{0.01cm} 
     Shaoteng Liu$^{2}$
     \hspace{0.01cm} 
    Haitian Zheng$^{2}$
     \hspace{0.01cm} 
    Jason Kuen$^2$ 
    \hspace{0.01cm} 
    Yuehuan Wang$^1$ 
     \hspace{0.01cm}
    Changxin Gao$^1$ 
     \hspace{0.01cm}
     Nong Sang$^{1}$
      \vspace{2mm}
     \\
    $^1$Key Laboratory of Image Processing and Intelligent Control,\\   School of Artificial Intelligence and Automation, Huazhong University of Science and Technology\\
      $^2$Adobe Research
      \\
}

\begin{document}


\twocolumn[{
\maketitle
    \centering
    \vspace{-29pt}
    \includegraphics[width=0.98\textwidth]{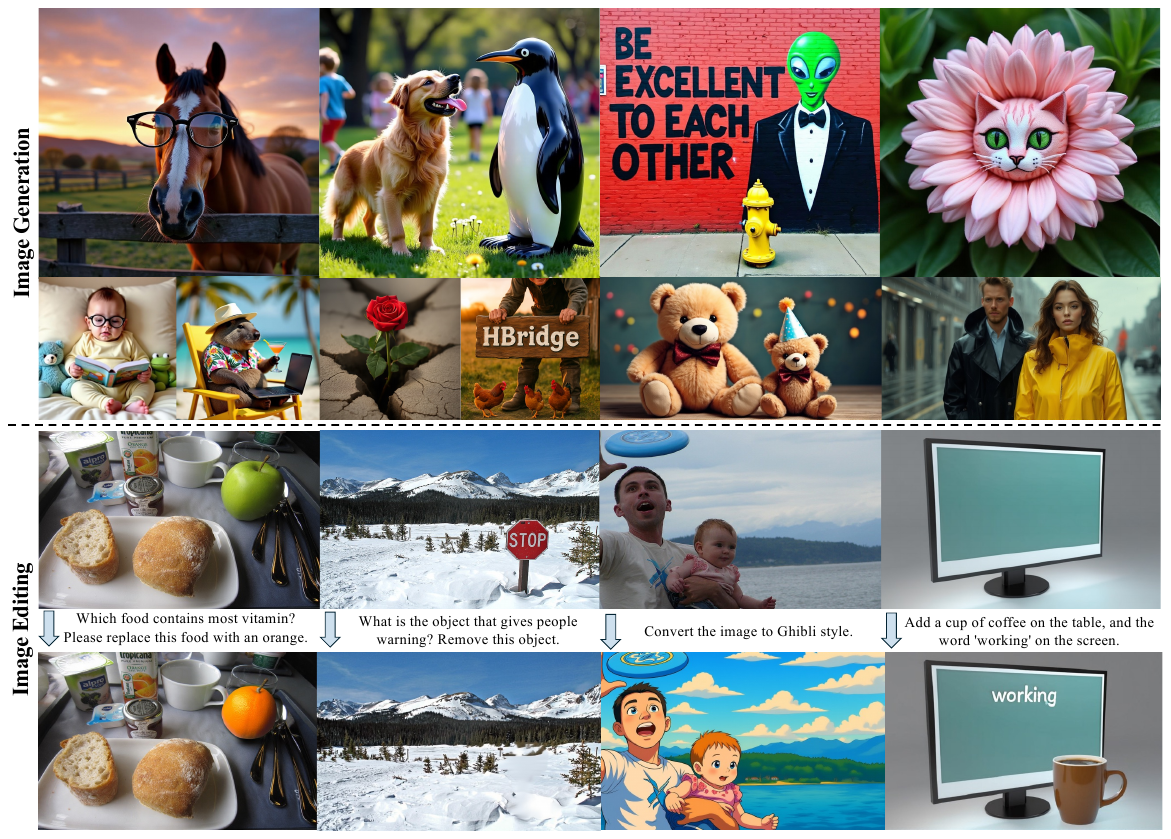}
    \vspace{-8pt}
    \captionof{figure}{
        Image generation and editing samples from \method, which achieves high-quality and photorealistic results.
    }
    \label{first_figure}
    \vspace{9pt}
    }
]

\begin{abstract}

\begin{figure*}[!htbp]
    \centering
    \includegraphics[width=0.99\linewidth]{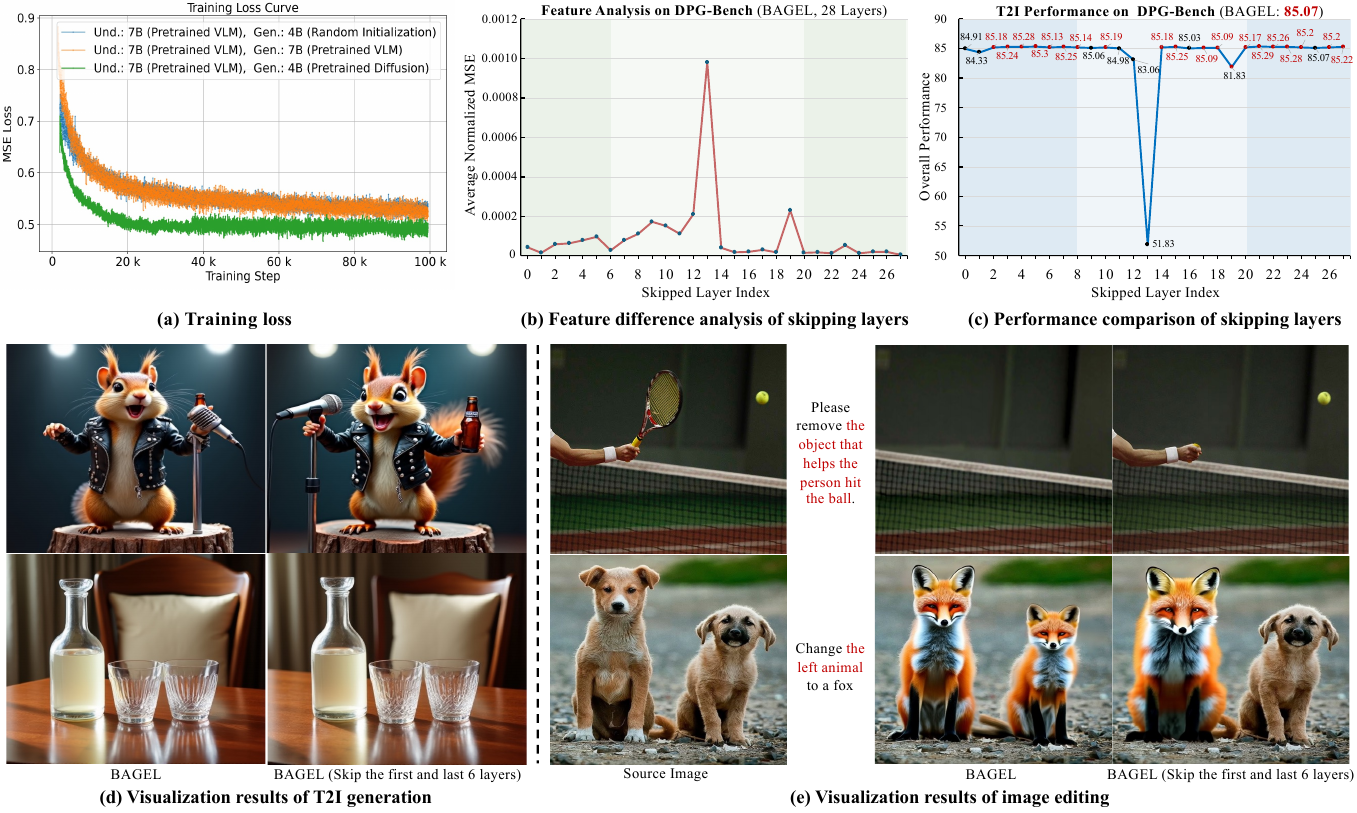}
    \vspace{-3mm}
    \caption{
    Motivation behind \method. The ``H'' refers to Heterogeneous (asymmetric) experts and the H-shape bridging that connects only the mid-layers between experts. Our asymmetric design is motivated by its faster convergence, as illustrated in (a). The symmetric 7B+7B baseline (\textcolor{orange}{orange}), initialized from Qwen2.5-VL-7B, converges slower than our asymmetric 7B+4B model (\textcolor{ForestGreen}{green}), where the 4B generative expert is initialized from OmniGen2. Remarkably, even a randomly initialized 7B+4B model (\textcolor{RoyalBlue}{blue}) converges at a similar speed to the well initialized 7B+7B model. The H-shape bridging is inspired by the correlation patterns in (b) and (c), which measure feature drift and T2I performance degradation when breaking individual cross-expert {self-attention} connections in BAGEL. These reveal that mid-layer connections dominate the performance, while shallow and deep-layer connections contribute minimally. By retaining only the essential mid-layer bridges, our architecture not only simplifies fusion but can outperform the original dense-fusion model. The corresponding visual examples in (d) and (e) further validate the above observation. Please note that results in (b)-(e) are derived directly from BAGEL.
%
    %
    }
    \label{fig:motivation}
    \vspace{-4mm}
\end{figure*}

Recent unified models integrate understanding experts (\eg, LLMs) with generative experts (\eg, diffusion models), achieving strong multimodal performance.
However,
recent advanced methods such as BAGEL and LMFusion follow the Mixture-of-Transformers (MoT) paradigm, adopting a symmetric design that mirrors one expert to another for convenient initialization and fusion, which remains suboptimal due to inherent modality discrepancies.
In this work, we
propose \method, an asymmetric H-shaped architecture that enables heterogeneous experts to optimally leverage pretrained priors from their respective modality domains.
Unlike prior dense fusion strategies that straightforwardly connect all layers between experts via shared attention, \method selectively bridges intermediate layers, reducing over 40\% attention sharing, which improves efficiency and enhances generation quality.
Shallow and deep layers, which capture modality-specific representations, are decoupled, while mid-layer bridging promotes semantic alignment.
To further strengthen cross-modal coherence, we introduce semantic reconstruction tokens that explicitly guide the generative expert to reconstruct 
visual semantic tokens of the target image.
%
Extensive experiments across multiple benchmarks demonstrate the effectiveness and superior performance of \method, establishing a new paradigm for unified multimodal generation.

\end{abstract}

\vspace{-10pt}
\section{Introduction}
\label{sec:intro}


Unified multimodal understanding and generation~\cite{wu2025omnigen2,bagel,Xfusion,xie2024showo,chen2025blip3,zhou2024transfusion,wu2025janus,team2024chameleon,gpt4o,tong2025metamorph,lu2025hyper,januspro,shi2024lmfusion,Qwenimage,wang2024emu3,cao2025hunyuanimage,xu2025tbac,li2025lavida,zhuang2025vargpt,huang2025ming,wu2024vila,wu2025harmonizing,wang2025unified,zhang2025nexus,zhang2025unified,wu2025openuni} has recently gained significant attention and emerged as a central direction for bridging understanding and generation within a unified architecture, powering multifunctional applications.
Current methods such as BAGEL~\cite{bagel}, 
BLIP3o~\cite{chen2025blip3},MetaQuery~\cite{metaqueries}, Mogao~\cite{mogao}, OmniGen2~\cite{wu2025omnigen2}, and LMFusion~\cite{shi2024lmfusion} exemplify this trend. 
They usually combine autoregressive large (vision-) language models (LLMs/VLMs)~\cite{qwen2.5vl,touvron2023llama} with a diffusion-based visual generator~\cite{stablediffusion,sd3}, where LLMs/VLMs are used for understanding tasks, and the diffusion generator focuses on visual synthesis.
Among these methods, the Mixture-of-Transformers (MoT) paradigm~\cite{bagel,mogao,shi2024lmfusion} employed by BAGEL and LMFusion demonstrates impressive capabilities and achieves state-of-the-art results across both understanding and generation tasks.
This paradigm uses a symmetric, densely connected design: it deploys two identical experts, both initialized from a pretrained LLM/VLM for understanding and generation, and bridges multimodal interactions through layer-by-layer shared self-attention.

Despite rapid progress, the unified MoT paradigm faces two fundamental limitations:

\noindent 1) \textbf{\textit{Symmetric architectures restrict generative priors}}.
The generative branch is typically initialized from an autoregressive LLM, since large-scale pretrained diffusion backbones with LLM-compatible architectures are not available. This mismatch prevents the generative expert from benefiting from strong pretrained priors, resulting in slow convergence (Fig.~\ref{fig:motivation}(a)) and high training cost, often comparable to random initialization.
In addition, understanding and generative models follow divergent scaling trends: language models now exceed 1T parameters~\cite{team2025kimi,li2025every,gpt4o}, while state-of-the-art generative models remain mostly below 20B~\cite{sd3,FLUX}.
These discrepancies indicate that heterogeneous experts are needed to fully exploit the strengths of existing pretrained models.

\noindent 2) \textbf{\textit{Dense layer-wise attention sharing ignores task asymmetry}}.
Understanding tasks usually rely on high-level semantic reasoning, whereas generation requires modeling fine-grained low-level structures. Sharing multimodal self-attention across all layers, including the earliest input layers and final output layers, can interfere with learning the task-specific feature spaces required by each branch.
Empirically, as shown in Fig.~\ref{fig:motivation} (b–e), the early and late layers of BAGEL~\cite{bagel} contribute minimally to final performance; skipping these connections does not degrade results and can even improve them.
Moreover, our analysis shows that dense layer-by-layer connections may encourage the generative expert to overfit shallow features from the understanding expert (Fig.~\ref{fig:Ablation_feature_analysis}), reducing its ability to capture high-level contextual semantics (Fig.~\ref{fig:Ablation_skip_layer}).

Motivated by these observations, we propose \method, an asymmetric H-shaped MoT architecture that unifies multimodal understanding and generation through heterogeneous experts and a mid-layer semantic bridge:
\begin{itemize}
    \item Unlike prior symmetric designs, \method pairs a large pretrained LLM with a diffusion-based generative expert, and replaces full-layer sharing with a selective mid-layer bridge, eliminating over 40\% of attention connections and preventing potential shallow overfitting.
    \item We introduce semantic reconstruction tokens that guide the generative branch to reconstruct ViT-level visual features, further enhancing semantic reasoning.
    \item Extensive experiments demonstrate that \method offers strong resource efficiency, attaining superior performance under lower training budgets. 
    %
    Remarkably, compared to BAGEL’s $\sim$2.5T T2I tokens, \method requires $\sim$200B T2I training tokens yet achieves even higher performance.
\end{itemize}

\vspace{-1mm}
\section{Related Work}
\label{sec:related_work}

\noindent
\textbf{Image Generation and Editing.} 
Diffusion models~\cite{sd3-medium,stablediffusion,chen2023pixart,podell2023sdxl,brooks2023instructpix2pix,controlnet,DAlle,dalle3,sd3,zhao2024ultraedit,ye2025imgedit,Emuedit,dit,qu2025tokenflow,liu2025step1x,mao2025ace,chen2025unireal} have become the dominant framework for plausible image generation and controllable editing.
Stable Diffusion~\cite{stablediffusion} introduces latent diffusion models, operating in a compressed latent space to achieve high-quality, efficient text-to-image synthesis. 
%
%
Recent advances in diffusion Transformers (DiT)~\cite{dit} have scaled visual generation to a new level. 
PixArt-$\alpha$~\cite{chen2023pixart} leverages cross-attention modules to inject text conditions into DiT backbone  for large-scale text-to-image training.
%
SD3~\cite{sd3} and FLUX~\cite{FLUX} introduce hybrid multimodal DiT blocks with cross-modal conditioning and apply flow matching~\cite{flowmatching} to optimize the model, significantly improving compositional semantic control and visual realism.
For image editing, typical methods such as InstructPix2Pix~\cite{brooks2023instructpix2pix}, ControlNet~\cite{controlnet} and Composer~\cite{huang2023composer} enable text-guided or structural control by fine-tuning conditional branches or adding spatial customized adapters.

\begin{figure*}[t]
    \centering
    \includegraphics[width=0.99\linewidth]{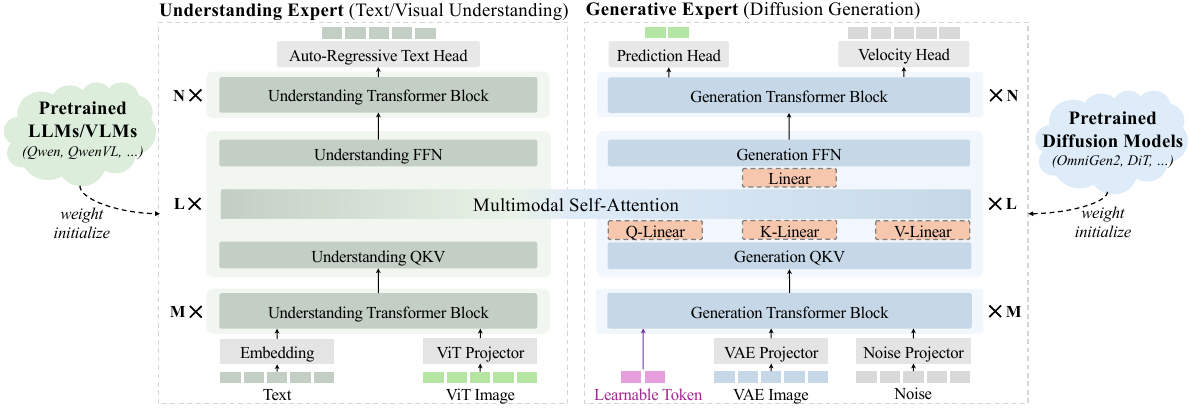}
    \vspace{-3mm}
    \caption{
    Overview of the proposed \method. We pair an arbitrary pretrained understanding expert with a pretrained generative expert and connect their mid-layers through self-attention. In practice, the understanding expert is typically a VLM, while the generative expert is often a DiT variant. QKV-Linear modules are applied to align their feature dimensions. 
    Additionally, we introduce learnable semantic tokens that explicitly reconstruct visual semantic tokens of the target image, improving text alignment and enhancing generation quality.
    }
    \label{fig:network}
    \vspace{-2mm}
\end{figure*}

\vspace{1mm}
\noindent\textbf{Unified Understanding and Generation.}
Unified understanding and generation models~\cite{bagel,lin2025uniworld,cao2025hunyuanimage,wang2024emu3,chen2025blip3,shi2024lmfusion,mogao} aim to jointly perform perception and generation within one framework. Two major paradigms have emerged: pure autoregressive (AR) and hybrid AR/diffusion architectures.
Pure autoregressive approaches~\cite{team2024chameleon,li2025onecat,jiao2025unitoken,qu2025tokenflow} treat multimodal data as token sequences, leveraging language modeling objectives for both understanding and generation.
%
Chameleon~\cite{team2024chameleon} integrates discrete visual tokens into a uniform
Transformer-based architecture and trains the whole model from scratch in an end-to-end manner.
%
%
UniToken~\cite{jiao2025unitoken} incorporates a combination
of discrete and continuous tokens to encode visual features and trains an auto-regressive generation
model for unified tasks.
However, recent AR models still struggle to achieve photorealistic synthesis due to the difficulty of unified tokenizers and the accumulation errors of autoregressive prediction.
%
The hybrid AR/diffusion paradigm~\cite{bagel,metaqueries,tong2025metamorph,wang2024emu3,zhou2024transfusion,xie2024showo,Showo2,cao2025hunyuanimage,mogao,shi2024lmfusion,wang2025ovis,ma2025janusflow} combines autoregressive LLM and diffusion model, and has attracted widespread attention.
MetaQuery~\cite{metaqueries} optimizes a set of learnable queries on the input side of the LLM and feeds the corresponding outputs into the diffusion model for generation.
Metamorph~\cite{tong2025metamorph} learns an LLM to predict continuous visual tokens, guiding the diffusion generator to synthesize visual content.
Recently, 
BAGEL~\cite{bagel} and Mogao~\cite{mogao} introduced a dense Mixture-of-Transformers (MoT) architecture that bridges the understanding and generative experts of identical Transformer architecture via fully shared attention, achieving state-of-the-art visual perception and synthesis performance.
%
%
%
While effective, they usually use symmetric experts with pretrained LLM initialization, 
overlooking the enormous potential of exploring powerful pretrained diffusion models to reduce resource consumption.
In contract, \method departs from these symmetric paradigms by introducing asymmetric experts with mid-layer semantic bridge, enabling the utilization of pretrained diffusion weights and preserving task-specific representation learning.
%

\section{HBridge}

We propose \method, a hybrid asymmetric MoT architecture that unifies multimodal understanding and generation within a single framework.
%
%
%
Without loss of generality,
we first provide a brief overview of \method.
%
Then, we will elaborate on the detailed mechanisms of each component.

\subsection{Overview}

Unlike prior symmetric MoT models~\cite{bagel,lu2025hyper,mogao,shi2024lmfusion} that use identical Transformer branches with fully shared attention, \method introduces heterogeneous experts for understanding and generation and connects them only through a mid-layer semantic bridge.
%
%
As shown in~\cref{fig:network}, the proposed \method introduces three key innovations:
(1) Heterogeneous  Experts, which allow the understanding and generation branches to adopt pretrained architectures better suited to their respective modalities. An understanding expert initialized from a large pretrained LLM or vision-language model, and a generative expert based on a diffusion-oriented Transformer;
(2) Mid-layer Semantic Bridge, which enables selective information exchange, and only a subset of middle layers is connected via cross-attention bridges, forming an H-shaped topology that balances independence and interaction;
(3) Semantic Reconstruction Tokens for explicit semantic grounding in generation.
%
Together, these components enable effective semantic–visual alignment while preserving the distinct inductive biases and pretrained priors of both branches.

%

%

\subsection{Heterogeneous Experts}

\textbf{Understanding Expert.}
The understanding branch is based on a large pretrained LLM/VLM~\cite{qwen2.5vl,team2024qwen2}.
It encodes text and image inputs into high-level conceptual representations and performs token-level autoregressive semantic reasoning.
This expert excels at contextual understanding and can serve as a semantic prior provider for visual generation.
We explore two variants, named Qwen2.5-VL-7B~\cite{qwen2.5vl} and Qwen2.5-0.5B~\cite{team2024qwen2}. To preserve the original powerful reasoning capability of the pretrained model, we freeze the weights of the understanding expert and focus on improving the visual generative ability of the generative expert.

\vspace{1mm}
\noindent
\textbf{Generative Expert.}
The generative expert adopts a full-attention DiT-style diffusion model from OmniGen2~\cite{wu2025omnigen2} pretrained for image synthesis.
Since the understanding and generative experts differ significantly in internal configurations, such as embedding dimension, normalization type, and attention head number, direct attention sharing is infeasible.
To resolve this, we introduce a QKV-Linear alignment module that enables the two experts to communicate within a unified latent space while preserving their own pretrained structures.
Specifically, given QKV features $U_{l}^{q}\in \mathbb{R} ^{L_{u}\times d_{u}^{q}}, U_{l}^{k}\in \mathbb{R} ^{L_{u}\times d_{u}^{k}}, U_{l}^{v}\in \mathbb{R} ^{L_{u}\times d_{u}^{v}} $ from the $l$-th layer of the understanding expert and $G_{l}^{q}\in \mathbb{R} ^{L_{g}\times d_{g}^{q}},G_{l}^{k}\in \mathbb{R} ^{L_{g}\times d_{g}^{k}},G_{l}^{v} \in \mathbb{R} ^{L_{g}\times d_{g}^{v}}$ from the $l$-th generative expert, where $L_{u}$ and $L_{g}$ are token length,  $d_{u}$ and $d_{g}$ mean the feature dimensions, we project the generation QKV features into a shared semantic space of understanding dimension:
\begin{equation}
    Q_{l} =W_{l}^{q}G_{l}^{q}, \quad K_{l} =W_{l}^{k}G_{l}^{k},\quad  V_{l} =W_{l}^{v}G_{l}^{v} 
\end{equation}
where $W_{l}^{q} \in \mathbb{R} ^{d_{g}^{q}\times d_{u}^{q}}, W_{l}^{k} \in \mathbb{R} ^{d_{g}^{k}\times d_{u}^{k}}, W_{l}^{v} \in \mathbb{R} ^{d_{g}^{v}\times d_{u}^{v}}$.
Cross-modal attention is then performed on the resulting features. After semantic information interaction, the outputs of the generation branch are projected back to the original space of the diffusion expert through a linear operation.
The diffusion model in~\cite{wu2025omnigen2}  contains 32 Transformer layers, higher than the 28 layers in Qwen2.5-VL-7B and the 24 layers in Qwen2.5-0.5B. We move the redundant layers of the generative expert into the Noise Projector to ensure multimodal interaction with the understanding expert.
To train the entire generative expert, we adopt the flow matching objective~\cite{flowmatching}  following previous practices~\cite{bagel,lin2025uniworld,mogao,shi2024lmfusion}.

\subsection{Mid-layer Semantic Bridge}

Full-layer attention sharing, as used in previous unified models~\cite{bagel,mogao,shi2024lmfusion}, may ignore intrinsic task asymmetry: understanding tasks require high-level semantic reasoning, while generation needs to capture low-level details. The shared attention in the early input layer and the late output layer may hinder the learning of the modality-specific representations related to each task. 
In addition, we find that the dense connected MoT architecture may result in potential overfitting to shallow textual semantics, bypassing high-level reasoning semantics.
Since many generative tasks, such as generating objects, can achieve promising results simply by utilizing shallow lexical or entity features from the frozen understanding expert during the training process, which causes the generative model to neglect extracting high-level reasoning cues.
To alleviate this, our \method introduces a mid-layer semantic bridge, which connects the two experts only within a selected range of $L$ intermediate layers, skipping the first $M$ and last $N$ connection layers.

\subsection{Semantic Reconstruction Tokens}

Generation tasks often require explicit semantic understanding, such as object relations, layout perception, and compositional reasoning.
To enhance this, we introduce Semantic Reconstruction Tokens (SRT) on the input side of the generative expert.
A small set of learnable tokens 
is appended to the generation input sequence.
In the experiment, we apply 16 learnable tokens.
During training, these tokens are supervised to reconstruct ViT-level semantic features of the ground-truth target image with cosine distance:
\begin{equation}
    \mathcal{L}_{SRT} = \mathrm{Distance_{cosine}}(\mathrm{Proj} (\mathrm{Token}_{SRT}^{out}) ,F_{\mathrm{ViT}})
\end{equation}
\begin{equation}
    \mathcal{L} = \mathcal{L}_{Flow matching} + \mathcal{L}_{SRT}
\end{equation}
where $F_{ViT}$ denotes adaptively pooled features extracted by a frozen pretrained ViT encoder of Qwen2.5-VL-7B to match the shape of the learnable tokens, and 
$\mathrm{Proj}(\cdot)$
is a lightweight projection head.
This auxiliary loss is jointly optimized together with the flow matching denoising loss.
This semantic reconstruction objective explicitly injects semantic supervision into the generative process, encouraging the model to internalize  relational semantics.

\section{Experiments}

In this section, we present a comprehensive quantitative and qualitative evaluation on various tasks to verify the effectiveness and superiority of the proposed \method.

\begin{table*}[t]
    \centering
    \small
    \caption{
    Quantitative comparison of text-to-image generation results with state-of-the-art methods on the DPG-Bench benchmark~\cite{DPGBench}.
    }
    \vspace{-3mm}
\setlength\tabcolsep{9.5pt}
    \begin{tabular}{lcccccc|c}
        \shline
        \hspace{-3mm}Method & \multicolumn{1}{c}{\# Params.} & \multicolumn{1}{c}{Global ($\uparrow$)} & \multicolumn{1}{c}{Entity ($\uparrow$)} & \multicolumn{1}{c}{Attribute ($\uparrow$)} & \multicolumn{1}{c}{Relation ($\uparrow$)} & \multicolumn{1}{c}{Other ($\uparrow$)} & 
        \multicolumn{1}{|c}
        {\textbf{Overall ($\uparrow$)}} \\
        \shline
        \multicolumn{8}{c}{\textit{\textbf{Diffusion models / Autoregressive models}}}  \\
            \hline
        \hspace{-3mm}LUMINA-Next~\cite{zhuo2024lumina} & 2B & 82.82 & 88.65 & 86.44 & 80.53 & 81.82 & 74.63 \\
        \hspace{-3mm}SDXL~\cite{podell2023sdxl}  & 2.6B  & 83.27 & 82.43 & 80.91 & 86.76 & 80.41 & 74.65 \\ 
        \hspace{-3mm}PlayGroundv2.5~\cite{li2024playground}  & - & 83.06 & 82.59 & 81.20 & 84.08 & 83.50 & 75.47 \\
        \hspace{-3mm}Hunyuan-DiT~\cite{li2024hunyuan}  & 1.5B & 84.59 & 80.59 & 88.01 & 74.36 & 86.41 & 78.87 \\
        \hspace{-3mm}PixArt-$\alpha$~\cite{chen2023pixart}  & 0.6B & 74.97 & - & - & 82.57 & - & 71.11 \\
        \hspace{-3mm}DALLE-3~\cite{dalle3}  & - & 90.97 & 89.61 & 88.39 & 90.58 & 89.83 & 83.50 \\
        \hspace{-3mm}SD3-medium~\cite{sd3-medium}  & 2B & 87.90 & 91.01 & 88.83 & 80.70 & 88.68 & 84.08 \\
        \hspace{-3mm}FLUX.1-dev~\cite{FLUX}  & 12B & 82.1 & 89.5 & 88.7 & 91.1 & 89.4 & 84.0 \\ 
        \hspace{-3mm}OmniGen~\cite{xiao2025omnigen}  & 3.8B & 87.90 & 88.97 & 88.47 & 87.95 & 83.56 & 81.16 \\
        \hspace{-3mm}Infinity~\cite{han2025infinity}  & 2B & 93.11 & - & - & 90.76 & - & 83.46 \\
        \hspace{-3mm}SimpleAR~\cite{wang2025simplear}  & 1.5B & 87.97 & - & - & 88.33 & - & 81.97 \\
        \hline
        \multicolumn{8}{c}{\textit{\textbf{Unified understanding and generation}}}  \\
            \hline
        \hspace{-3mm}Show-o~\cite{xie2024showo}  & 1.3B & 79.33 & 75.44 & 78.02 & 84.45 & 60.80 & 67.27 \\
        \hspace{-3mm}EMU3~\cite{wang2024emu3}   & 8.5B & 85.21 & 86.68 & 86.84 & 90.22 & 83.15 & 80.60 \\
        \hspace{-3mm}TokenFlow-XL~\cite{qu2025tokenflow} & 14B & 78.72 & 79.22 & 81.29 & 85.22 & 71.20 & 73.38 \\ 
        \hspace{-3mm}Janus~\cite{wu2025janus} & 1.5B & 82.33 & 87.38 & 87.70 & 85.46 & 86.41 & 79.68 \\
        \hspace{-3mm}Janus Pro~\cite{januspro} & 7B & 86.90 & 88.90 & 89.40 & 89.32 & 89.48 & {84.19} \\
        \hspace{-3mm}BLIP3-o 4B~\cite{chen2025blip3} & 3B + 1.4B & - & - & - & - & - & 79.36 \\
        \hspace{-3mm}BLIP3-o 8B~\cite{chen2025blip3} & 7B + 1.4B & - & - & - & - & - & 81.60 \\
        %
        \hspace{-3mm}UniWorld-V1~\cite{lin2025uniworld} & 7B + 12B & 83.64 & 88.39 & 88.44 & 89.27 & 87.22 & 81.38 \\
        \hspace{-3mm}{OmniGen2}~\cite{wu2025omnigen2}  & 3B + 4B & 88.81 & 88.83 & 90.18 & 89.37 & 90.27 & 83.57 \\
        \hspace{-3mm}BAGEL~\cite{bagel} & 7B + 7B & 88.94 & 90.37 & 91.29 & 90.82 & 88.67 & \underline{85.07} \\
        \rowcolor{myblue}\hspace{-3mm}\textbf{\method} & 7B + 4B & 91.78 & 91.82 & 90.23 & 90.06 & 88.42 & \textbf{85.23} \\
        \shline
    \end{tabular}
\label{tab:dpgbench}
\end{table*}

\begin{table*}[t]
    \centering
\small

    \caption{
    Text-to-image results on  GenEval benchmark~\cite{ghosh2023geneval}. 
    ``${*}$" means reproducing the result using the official open-source checkpoints.
    }
    \vspace{-3mm}
\setlength\tabcolsep{4.5pt}
    {
        \begin{tabular}{lcccccc|c}
            \shline
            \hspace{-1.2mm}Method & \multicolumn{1}{c}{Single object ($\uparrow$)} & \multicolumn{1}{c}{Two object ($\uparrow$)} & \multicolumn{1}{c}{Counting ($\uparrow$)} & \multicolumn{1}{c}{Colors ($\uparrow$)} & \multicolumn{1}{c}{Position ($\uparrow$)} & \multicolumn{1}{c}{Color attri. ($\uparrow$)} & \multicolumn{1}{|c}{\textbf{Overall ($\uparrow$)}} \\
            \shline
            \multicolumn{8}{c}{\textit{\textbf{Diffusion models / Autoregressive models}}}  \\
            \hline
            \hspace{-1.2mm}LUMINA-Next~\cite{zhuo2024lumina} & 0.92 & 0.46 & 0.48 & 0.70 & 0.09 & 0.13 & 0.46 \\
            \hspace{-1.2mm}SD3-medium~\cite{sd3-medium} & 0.99 & 0.94 & 0.72 & 0.89 & 0.33 & 0.60 & 0.74 \\
            \hspace{-1.2mm}FLUX.1-dev~\cite{FLUX} & 0.99 & 0.81 & 0.79 & 0.74 & 0.20 & 0.47 & 0.67 \\
            \hspace{-1.2mm}NOVA~\cite{deng2024autoregressive} & 0.99 & 0.91 & 0.62 & 0.85 & 0.33 & 0.56 & 0.71 \\
            \hspace{-1.2mm}OmniGen~\cite{xiao2025omnigen} & 0.98 & 0.84 & 0.66 & 0.74 & 0.40 & 0.43 & 0.68 \\ 
            \hspace{-1.2mm}Infinity~\cite{han2025infinity} & - & 0.85 & - & - &  0.49 & 0.57  & 0.73 \\ 
            \hspace{-1.2mm}SimpleAR~\cite{wang2025simplear} & - & 0.90 & - & - & 0.28 & 0.45 & 0.63 \\ 
            \hline
            \multicolumn{8}{c}{\textit{\textbf{Unified understanding and generation}}}  \\
            \hline
            \hspace{-1.2mm}Chameleon~\cite{team2024chameleon} & - & - & - & - & - & - & 0.39 \\ 
            \hspace{-1.2mm}TokenFlow-XL~\cite{qu2025tokenflow} & 0.95 & 0.60 & 0.41 & 0.81 & 0.16 & 0.24 & 0.55 \\ 
            \hspace{-1.2mm}Janus~\cite{wu2025janus} & 0.97 & 0.68 & 0.30 & 0.84 & 0.46 & 0.42 & 0.61 \\
            \hspace{-1.2mm}ILLUME~\cite{wang2025illume} & 0.99 & 0.86 & 0.45 & 0.71 & 0.39 & 0.28 & 0.61 \\
            \hspace{-1.2mm}Transfusion~\cite{zhou2024transfusion} & - & - & - & - & - & - & 0.63 \\
            \hspace{-1.2mm}Janus Pro~\cite{januspro} & 0.99 & 0.89 & 0.59 & 0.90 & 0.79 & 0.66 & {0.80} \\
            \hspace{-1.2mm}Show-o~\cite{xie2024showo} & 0.98 & 0.80 & 0.66 & 0.84 & 0.31 & 0.50 & 0.68 \\
            %
            \hspace{-1.2mm}UniWorld-V1~\cite{lin2025uniworld} & 0.99 & 0.93 & 0.79 & 0.89 & 0.49 & 0.70 & {0.80} \\
            \hspace{-1.2mm}OmniGen2~\cite{wu2025omnigen2} & 1 & 0.95 & 0.64 & 0.88 & 0.55 & 0.76 & {0.80} \\   
            \hspace{-1.2mm}Nexus-Gen(7B+12B)~\cite{zhang2025nexus} & 0.97 & 0.93 & 0.64  & 0.88  & 0.83  & 0.62 & \underline{0.81} \\
            \hspace{-1.2mm}BAGEL$^{*}$~\cite{bagel} & 1 & 0.94 & 0.79 & 0.88 & 0.55 & 0.66 & {0.80} \\
            \rowcolor{myblue}\hspace{-1.2mm}\textbf{\method} & 0.97 & 0.94 & 0.73 & 0.93 & 0.63 & 0.78 & \textbf{0.83} \\
            %
            \hline
            \multicolumn{8}{c}{\textit{\textbf{Unified understanding and generation with LLM re-writer}}}  \\
            \hline
            \hspace{-1.2mm}$\text{Emu3-Gen}$~\cite{wang2024emu3} & 0.99 & 0.81 & 0.42 & 0.80 & 0.49 & 0.45 & 0.66 \\\hspace{-1.2mm}$\text{MetaQuery-XL}$~\cite{metaqueries} & - & - & - & - & - & - & 0.80 \\
            \hspace{-1.2mm}$\text{BLIP3-o}$ 4B~\cite{chen2025blip3} & - & - & - & - & - & - & 0.81 \\
            \hspace{-1.2mm}$\text{BLIP3-o}$ 8B~\cite{chen2025blip3} & - & - & - & - & - & - & 0.84 \\

            \hspace{-1.2mm}$\text{UniWorld-V1}$~\cite{lin2025uniworld} & 0.98 & 0.93 & 0.81 & 0.89 & 0.74 & 0.71 & 0.84 \\
            \hspace{-1.2mm}{OmniGen2}~\cite{wu2025omnigen2} & 0.99 & 0.96 & 0.74 & 0.98 & 0.71 & 0.75 & \underline{0.86} \\
            \hspace{-1.2mm}$\text{BAGEL}^{*}$~\cite{bagel} & 0.98 & 0.95 & 0.81 & 0.92 & 0.73 & 0.74 & \underline{0.86} \\
            \hspace{-1.2mm}Hyper-BAGEL~\cite{lu2025hyper} & 0.99 & 0.94 & 0.86 & 0.94 & 0.72 & 0.74 & \underline{0.86} \\
            \rowcolor{myblue}\hspace{-1.2mm}\textbf{\method} & 1 & 0.96 & 0.80 & 0.94 & 0.77 & 0.78 & \textbf{0.87} \\
            \shline
        \end{tabular}
    }
    \label{tab:geneval}
\end{table*}

\subsection{Experimental Setup}


\noindent \textbf{Implementation Details.}
Our method involves two pretrained experts.
The understanding expert is initialized from a pretrained LLM/VLM backbone. 
Two variants, \ie Qwen2.5-0.5B~\cite{team2024qwen2} and Qwen2.5-VL-7B~\cite{qwen2.5vl}, are adopted to verify the effectiveness and implantability of the proposed method.
The generative expert employs a 4B DiT from OminGen2~\cite{wu2025omnigen2}, and we add some linear layers to map the original dimension of the input tokens to match the attention operation, resulting in about 4B parameters.
We skip the first of 6 layers and the last 6 layers, and 
only the middle layers are bridged for multimodal attention exchange. 
We use AdamW optimizer with a consistent learning rate of 1e-4 to optimize our model.
The total training step is about 200k.
%
All experiments are implemented in PyTorch with mixed precision on 64 H100/A100/A800 GPUs. 
Some experiments are trained on 16 GPUs, and we use the gradient accumulation strategy to approximate training on 64 GPUs.
We collected approximately 400M images from external open-source datasets~\cite{chen2025blip3,lin2025uniworld,wu2025omnigen2,zhao2024ultraedit} and internal databases to train the model.

\noindent \textbf{Benchmarks and Metrics.}
%
For multimodal understanding, 
since the understanding expert is frozen during the training process, the capability is preserved.
For visual generation, we evaluate \method on a diverse set of generative benchmarks that cover both image generation (\ie,  DPG-Bench~\cite{DPGBench} and  GenEval~\cite{ghosh2023geneval}) and image editing (\ie, ImgEdit-Bench~\cite{ye2025imgedit}).
The benchmark metrics include counting, relationships, semantic understanding, \etc, and the final average result is used to indicate the final performance.
Unless otherwise specified, our default setting is to use a \textbf{7B+4B} setting (understanding expert: Qwen2.5-VL-7B~\cite{qwen2.5vl}, generative expert: 4B DiT model~\cite{wu2025omnigen2}) for comparison with existing state-of-the-art methods. To effectively verify the effectiveness of our method and save computing resources, we adopt a \textbf{0.5B+4B} setting (understanding expert: Qwen2.5-0.5B~\cite{team2024qwen2}, generative expert: 4B DiT model~\cite{wu2025omnigen2}) for ablation studies.

\begin{table*}[t]
    \centering
    \small
    \caption{Quantitative results of image editing on ImgEdit-Bench~\cite{ye2025imgedit}. ``Overall" indicates the average score across all editing sub-tasks.
    %
    }
    \vspace{-3mm}
\setlength\tabcolsep{7.0pt}
    
        \begin{tabular}{lccccccccc|c}
            \shline
            \hspace{-1.7mm}\textbf{Method} & Add & Adjust & Extract & Replace & Remove & Background & Style & Hybrid & Action & \textbf{Overall ($\uparrow$)} \\
            \shline
            \hspace{-1.7mm}MagicBrush~\cite{zhang2024magicbrush} & 2.84 & 1.58 & 1.51 & 1.97 & 1.58 & 1.75 & 2.38 & 1.62 & 1.22 & 1.90 \\
            \hspace{-1.7mm}Instruct-P2P~\cite{brooks2023instructpix2pix} & 2.45 & 1.83 & 1.44 & 2.01 & 1.50 & 1.44 & 3.55 & 1.2 & 1.46 & 1.88 \\
            
            \hspace{-1.7mm}AnyEdit~\cite{yu2025anyedit} & 3.18 & 2.95 & 1.88 & 2.47 & 2.23 & 2.24 & 2.85 & 1.56 & 2.65 & 2.45 \\
            \hspace{-1.7mm}UltraEdit~\cite{zhao2024ultraedit} & 3.44 & 2.81 & {2.13} & 2.96 & 1.45 & 2.83 & 3.76 & 1.91 & 2.98 & 2.7 \\
            \hspace{-1.7mm}Step1X-Edit~\cite{liu2025step1x} & {3.88} & 3.14 & 1.76 & 3.40 & 2.41 & 3.16 & {4.63} & {2.64} & 2.52 & 3.06 \\
            \hspace{-1.7mm}ICEdit~\cite{zhang2025context} & 3.58 & {3.39} & 1.73 & 3.15 & 2.93 & 3.08 & 3.84 & 2.04 & 3.68 & 3.05 \\
            \hspace{-1.7mm}UniWorld-V1~\cite{lin2025uniworld} & {3.82} & {3.64} & {2.27} & {3.47} & {3.24} & 2.99 & 4.21 & {2.96} & 2.74 & {3.26} \\
            \hspace{-1.7mm}{OmniGen2}~\cite{wu2025omnigen2} & 3.57 & 3.06 & 1.77 & {3.74} & {3.2} & {3.57} & {4.81} & 2.52 & {4.68} & \underline{3.44} \\
            \hspace{-1.7mm}BAGEL~\cite{bagel} & 3.56 & 3.31 & 1.7 & 3.3 & 2.62 & {3.24} & 4.49 & 2.38 & {4.17} & 3.20 \\
            \rowcolor{myblue}\hspace{-1.7mm}\textbf{\method} & 3.91 & 3.23 & 2.13 & 3.79 & 3.21 & 3.50 & 4.31 & 3.44 & 4.32 & \textbf{3.54} \\
            \shline
        \end{tabular}
    \label{tab:imgedit}
    \vspace{-1.6mm}
\end{table*}

\begin{table}[t]
    \centering
    \small
    \caption{Ablation study on the number of skipped layers under the \textbf{0.5B+4B} setting.  LLM re-writer is not used on GenEval.
    }
    \vspace{-3mm}
    \setlength\tabcolsep{22.0pt}
    \begin{tabular}{l|c|c}
    \shline
      \hspace{-7mm}Setting   &  DPG-Bench & GenEval \\
      \shline
       \hspace{-7mm}M=0, \hspace{2mm} N=0  & 78.19 & 0.61 \\
       \hspace{-7mm}M=2, \hspace{2mm} N=2  & 78.63  & 0.62 \\
       \hspace{-7mm}M=4, \hspace{2mm} N=4  & 79.20 &  \textbf{0.66} \\
       \hspace{-7mm}M=0, \hspace{2mm} N=6  & 79.36 & 0.63 \\
       \hspace{-7mm}M=6, \hspace{2mm} N=6  & \textbf{80.03} & \textbf{0.66} \\
       \hspace{-7mm}M=6, \hspace{2mm} N=0  & 79.26 & 0.64 \\
       \hspace{-7mm}M=8, \hspace{2mm} N=8  & 73.51 & 0.54 \\
       \hspace{-7mm}M=10, \hspace{0.3mm} N=10  & 69.05 & 0.35 \\
       \shline
    \end{tabular}
    \label{tab:ablation_mn}
    \vspace{-2mm}
\end{table}

\subsection{Comparison with State-of-the-Arts}

\noindent
\textbf{Multimodal Visual Understanding.} 
Our method leverages the Qwen2.5-VL-7B model as the understanding expert under the 7B+4B setting, which is frozen during training to fully preserve its native multimodal understanding capabilities.
Like previous methods~\cite{lin2025uniworld}, our model can achieve solid scores of 83.5 on MMBench~\cite{liu2024mmbench}, 58.6 on MMMU~\cite{yue2024mmmu}, and 67.1 on MM-Vet~\cite{yu2023mm}, demonstrating strong visual-language reasoning performance.

\vspace{1mm}
\noindent
\textbf{Evaluation on Text-to-Image Generation.} 
%
We evaluate the performance of text-to-image generation on two widely-used benchmarks, named DPG-Bench~\cite{DPGBench} and GenEval~\cite{ghosh2023geneval}. 
The results are shown in~\cref{tab:dpgbench} and~\cref{tab:geneval}.
From the results, we can observe that our method achieves state-of-the-art results over competitive counterparts, such as UniWorld-V1, OmniGen2, and BAGEL.
Despite with only 4B generation parameters, \method can outperform BAGEL (7B+7B) and UniWorld-V1 (7B+12B), demonstrating the superiority and effectiveness of \method.
It's worth noting that
compared to BAGEL trained with $\sim$2.5T T2I tokens, the proposed \method requires only $\sim$200B T2I tokens to achieve even better results.

\vspace{1mm}
\noindent
\textbf{Evaluation on Image Editing.}
As shown in~\cref{tab:imgedit}, we evaluate the image editing performance of our method on ImgEdit-Bench~\cite{ye2025imgedit}. 
The results indicate that \method achieves the best overall performance over all competitors, such as BAGEL, OmniGen2 and UniWorld-V1.

\begin{figure}
    \centering
    \includegraphics[width=0.99\linewidth]{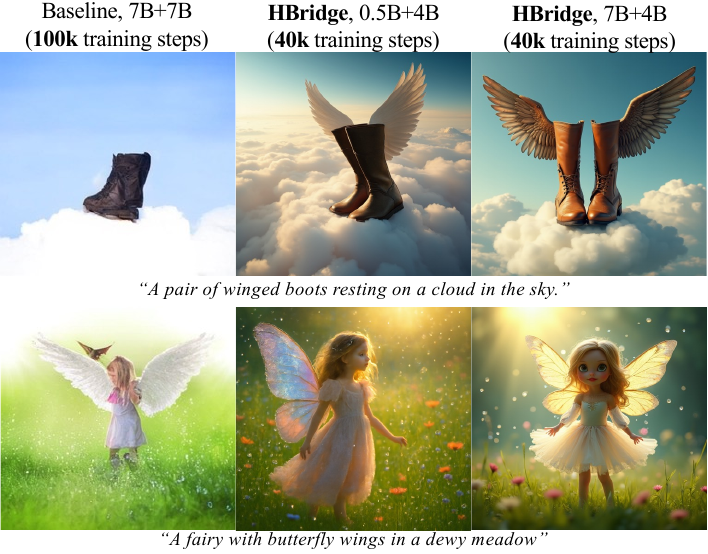}
    \vspace{-3mm}
    \caption{Ablation study on different initialization manners.
    We treat the understanding and generative experts initialized with Qwen2.5-VL-7B as the baseline, and our \method utilizes the pretrained diffusion initialization from~\cite{wu2025omnigen2}.
    }
    \label{fig:Ablation_initialization}
    \vspace{-2mm}
\end{figure}

\begin{figure}
    \centering
    \includegraphics[width=0.99\linewidth]{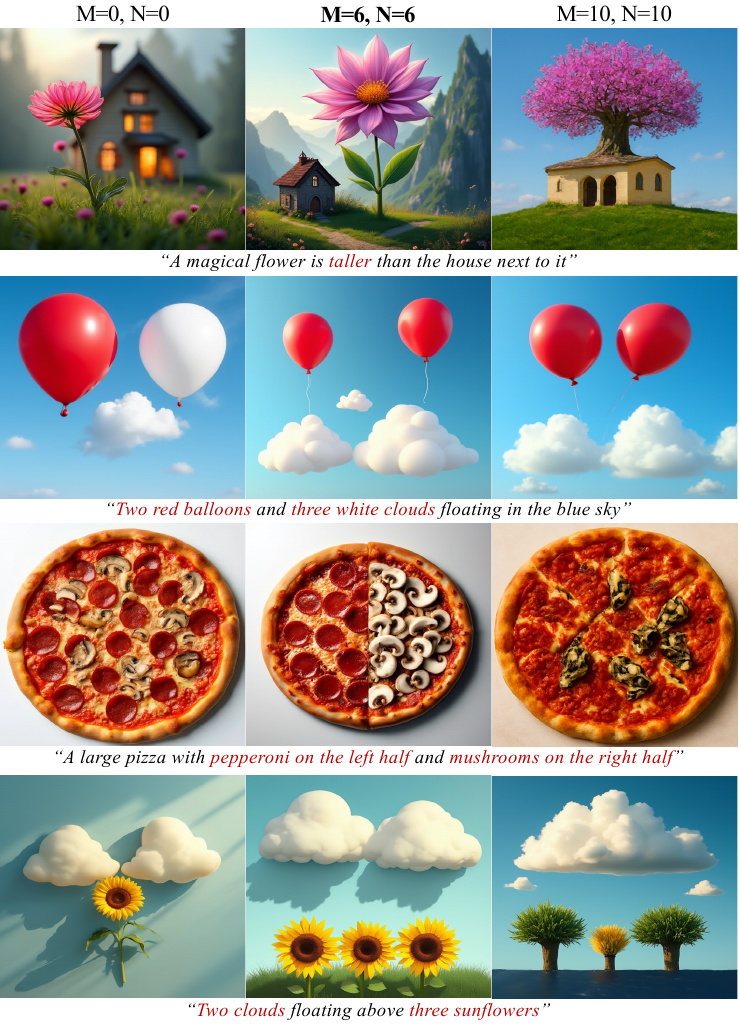}
    \vspace{-3mm}
    \caption{Ablation study on varying the number of skipped layers.
    }
    \label{fig:Ablation_skip_layer}
    \vspace{-2mm}
\end{figure}

\begin{figure}
    \centering
    \includegraphics[width=0.99\linewidth]{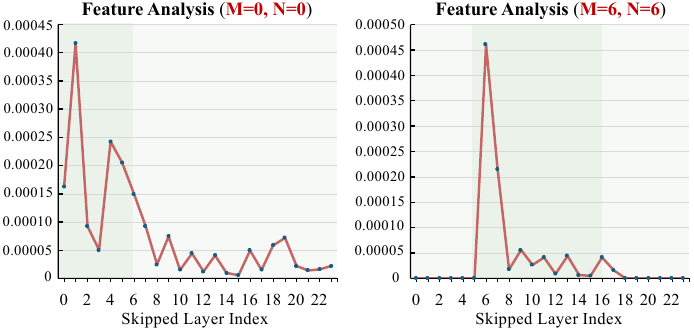}
    \vspace{-4mm}
    \caption{Analysis of varying the skipped layer on DPG-Bench.}
    \label{fig:Ablation_feature_analysis}
    \vspace{-2mm}
\end{figure}

\begin{table}[t]
    \centering
    \small
    \caption{
    Ablation study on semantic reconstruction tokens. 
    }
    \vspace{-3mm}
    \setlength\tabcolsep{12pt}
    \begin{tabular}{l|c|c}
    \shline
      \hspace{-4mm}Setting   &  DPG-Bench & GenEval \\
      \shline
      \hspace{-4mm}w/o semantic tokens  & 79.57 & 0.65 \\
       \hspace{-4mm}\textbf{w/ semantic tokens (Ours)}  & \textbf{80.03} & \textbf{0.66}
        \\
       \shline
    \end{tabular}
    \label{tab:ablation_semantic_token}
\end{table}

\begin{table}[t]
    \centering
    \small
    \caption{
    Ablation study on different cross-expert fusion manners. 
    }
    \vspace{-3mm}
    \setlength\tabcolsep{17.5pt}
    \begin{tabular}{l|c|c}
    \shline
      \hspace{-6mm}Setting   &  DPG-Bench & GenEval \\
      \shline
      \hspace{-6mm}Shallow fusion  & 74.53  & 0.58  \\
       \hspace{-6mm}\textbf{Deep fusion (Ours)}  & \textbf{80.03} & \textbf{0.66}
        \\
       \shline
    \end{tabular}
    \label{tab:ablation_shallow_deep}
\end{table}

\subsection{Ablation Study}

We conduct ablation studies to validate the rationality of each component.
Please refer to \textbf{\textit{appendix}} for more results.

\vspace{1mm}
\noindent
\textbf{Effect of Pretrained Diffusion Expert.} 
As displayed in~\cref{fig:Ablation_initialization}, \method training with merely 40k steps can produce high-fidelity images.
While replacing the pretrained diffusion backbone with a VLM initialized DiT leads to a sharp drop in generation quality even with more training steps.
This confirms that our asymmetric heterogeneous expert design effectively leverages the pretrained diffusion priors while allowing flexible multimodal integration through QKV-Linear alignment.
In addition, we can also observe that larger understanding experts (7B vs. 0.5B) can achieve better results with high visual quality.
%

\vspace{1mm}
\noindent
\textbf{Effect of Mid-layer Semantic Bridge.} 
We conduct an ablation study on different values of $M$ and $N$ to verify the effectiveness of the mid-layer bridge. 
The quantitative comparisons are shown in~\cref{tab:ablation_mn}, and M=N=6 achieves the best results on both DPG-Bench and GenEval.
%
%
%
%
%
In~\cref{fig:Ablation_skip_layer}, we display some visualization cases and can observe that
when M=N=10, the semantics of some objects may be overlooked, such as ``mushroom" and ``sunflower". We attribute this to the fact that there are too few semantic injection layers for visual-text alignment, making it difficult to learn reliable prompt following.
When M=N=0, the results remain visually realistic but may fail to match certain attributes, relations, and numerical terms, such as ``taller", ``three", etc.
In~\cref{fig:Ablation_feature_analysis}, we can notice that when M=N=0, there is an overfitting phenomenon here, where the shallow semantic layer dominates the generation process, which makes it difficult to understand some complex prompts, since complex semantic understanding is generally in the middle and deep layers of a large understanding model~\cite{ju2024large,fan2024not}.
During the training process,
many generative tasks, such as generating objects, can achieve promising results simply by utilizing shallow lexical or entity features from the frozen understanding expert, which causes the generative model to neglect extracting high-level reasoning cues.
Among these, M=N=6 achieves a good balance.

\vspace{1mm}
\noindent
\textbf{Effect of Semantic Reconstruction Tokens.}
%
As illustrated
in~\cref{tab:ablation_semantic_token}, the proposed semantic reconstruction tokens help to improve the performance.
We attribute to that
the learnable tokens explicitly inject semantic supervision by reconstructing ViT features, enforcing a stronger link between textual semantics and visual generation.

\vspace{1mm}
\noindent
\textbf{Shallow Fusion vs. Deep Fusion.}
Many recent advanced methods~\cite{lin2025uniworld,Qwenimage} use the output embeddings of the last layer in LLMs/VLMs to guide subsequent diffusion model.
We further compare this shallow fusion strategy with our deep fusion manner. The comparison is exhibited in~\cref{tab:ablation_shallow_deep}.
%
The deep fusion manner achieves better performance, and we believe the reason is that deep fusion can utilize rich, multi-layer semantic information from the understanding expert, while shallow fusion can only utilize the last layer.

\vspace{1mm}
\noindent
\textbf{Pluggability.}
%
As shown in~\cref{tab:ablation_on_bagel}, we insert the proposed mid-layer semantic bridge and
semantic reconstruction tokens into the open-source pretrained BAGEL and find that performance can be further improved, validating the insertability and generalization of the proposed components.

\begin{table}[t]
    \centering
    \small
    \caption{Generalization experiments based on BAGEL (7B+7B). ``MSB" and ``SRT" are abbreviations for Mid-layer Semantic Bridge and Semantic Reconstruction Tokens, respectively. ``$\dagger$" refers to the finetuned method using our collected data. 
    }
    \vspace{-3mm}
    \setlength\tabcolsep{2.5pt}
    \begin{tabular}{l|c|c|c}
    \shline
      \hspace{-0.5mm}Method   &  DPG-Bench & GenEval & ImgEdit-Bench \\
      \shline
      \hspace{-0.5mm}BAGEL~\cite{bagel}  & 85.07  & 0.80 & 3.20  \\
      \hspace{-0.5mm}BAGEL$^{\dagger}$  & 85.20  & 0.82 & 3.25  \\
       \hspace{-0.5mm}BAGEL + MSB   & {85.41} & \textbf{0.84} & 3.34 
        \\
        \hspace{-0.5mm}\textbf{BAGEL + MSB + SRT}   & \textbf{85.55} & \textbf{0.84} & \textbf{3.38} 
        \\
       \shline
    \end{tabular}
    \label{tab:ablation_on_bagel}
    \vspace{-2mm}
\end{table}
\section{Conclusion and Limitations}

In this work, we presented \method, a unified H-shaped framework that integrates multimodal understanding and generation through an asymmetric MoT architecture.
By introducing heterogeneous experts, a mid-layer semantic bridge, and semantic reconstruction tokens, \method achieves superior performance with low training cost.
%
%

\vspace{1mm}
\noindent
\textbf{Limitations.} 
Despite its effectiveness, \method still faces several limitations: 1)
to reduce experimental costs,
we usually assume that M=N and that both are even numbers in our experiments, and more detailed settings for different combinations of M and N are not explored;
2) 
the mid-layer bridge currently uses fixed coupling layers. Dynamic layer selection based on semantic salience may further improve performance and flexibility in the future.



\vspace{-0.1mm}

{
    \small
    \bibliographystyle{ieeenat_fullname}
    \bibliography{main}
}

\clearpage


\twocolumn[
{
        \centering
        \Large
        \textbf{\thetitle}\\
        \vspace{0.3em}Supplementary Material \\
        \vspace{1.5em}
    \centering
    \vspace{1pt}
    \includegraphics[width=0.99\textwidth]{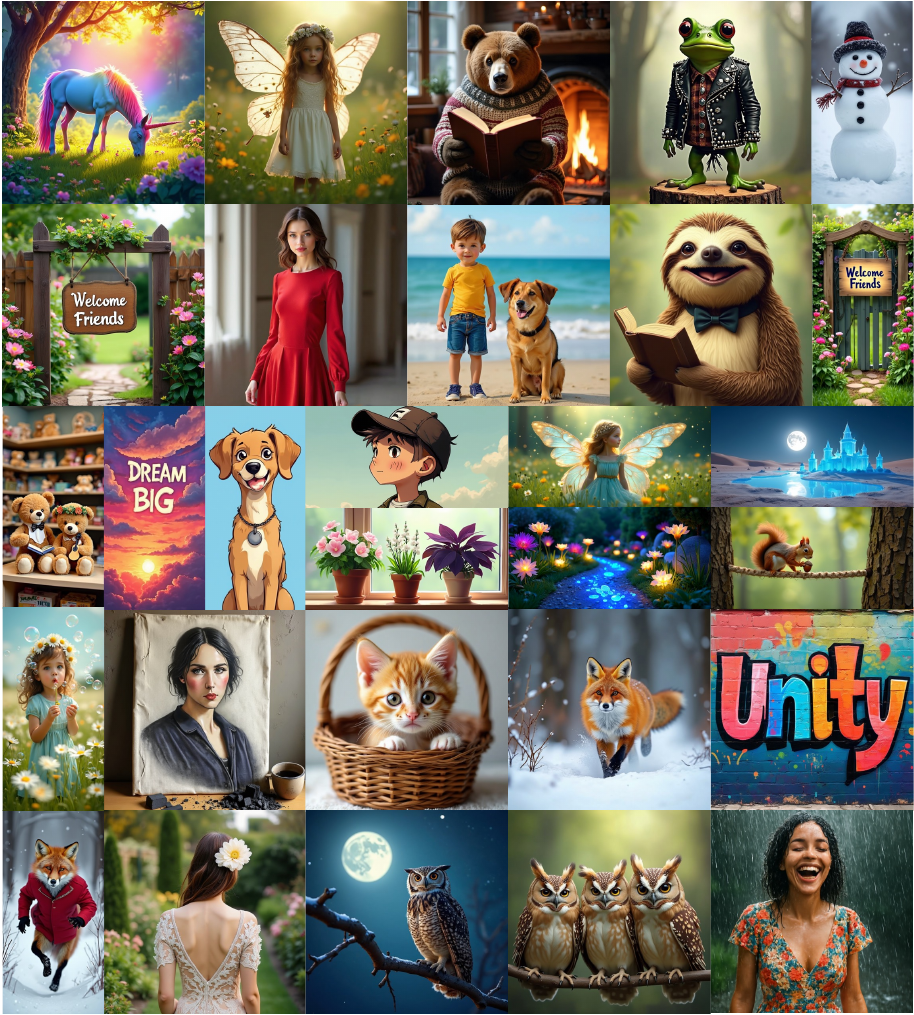}
    \vspace{-8pt}
    \captionof{figure}{
        More qualitative results of text-to-image generation synthesized by the proposed \method.
    }
    \label{fig:Supp_more_t2i}
    \vspace{18pt}
    }
]



\twocolumn[
{
    \centering
    \includegraphics[width=0.99\textwidth]{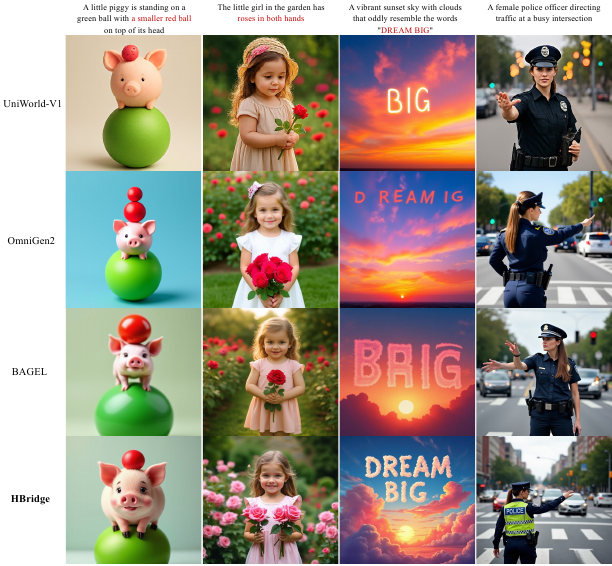}
    \vspace{-8pt}
    \captionof{figure}{
        Comparison with the state-of-the-art methods such as UniWorld-V1, OmniGen2 and BAGEL  on text-to-image generation task.
    }
    \label{fig:suppl_compare_editing_t2i}
    \vspace{15pt}
    }
]

Due to the page limit of the main document, we place some supplementary results and details in the appendix.

\begin{figure}[t]
    \centering
    \includegraphics[width=0.99\linewidth]{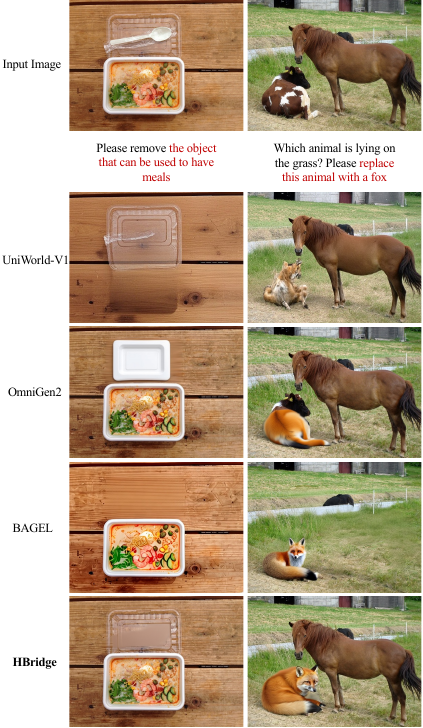}
    \vspace{-3mm}
    \caption{Comparison with state-of-the-art methods on image editing tasks, including object removal and replacement.}
    \label{fig:suppl_compare_editing_bagel}
\end{figure}

\begin{figure}[t]
    \centering
    \includegraphics[width=0.99\linewidth]{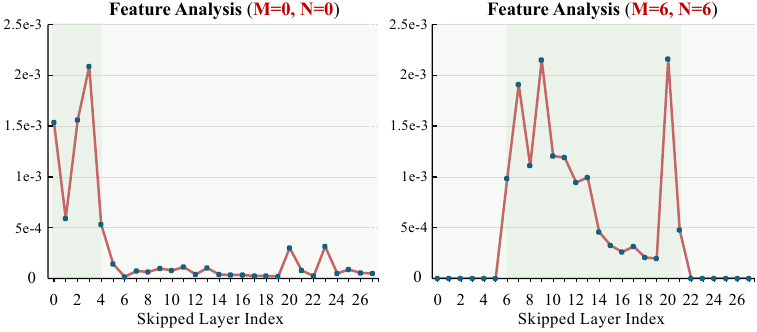}
    \vspace{-3mm}
    \caption{Analysis of varying the skipped layer under the \textbf{7B+4B} setting with 28 Transformer layers. We disconnect the multimodal self-attention layer by layer and analyze the differences in output features caused by disconnecting and reconnecting the multimodal self-attention layer. These differences are measured using the average normalized MSE~\cite{poli1993use} on DPG-Bench. }
    \label{fig:Supp_Ablation_feature_analysis}
\end{figure}

\section{More Qualitative Results}
\label{sec:Qualitative_Results}

As exhibited in~\cref{fig:Supp_more_t2i}, we show more high-quality, photorealistic text-to-image cases generated by \method with different resolution rates.
These examples demonstrate excellent spatial layout, quantity control, and text rendering, validating powerful generative capabilities of \method. 

In addition, we qualitatively compare our method with the state-of-the-art methods, including  UniWorld-V1~\cite{lin2025uniworld}, OmniGen2~\cite{wu2025omnigen2}, and BAGEL~\cite{bagel}. The text-to-image results are shown in ~\cref{fig:suppl_compare_editing_t2i}.
Our method demonstrates better semantic coherence and visual quality.
The editing results are displayed in~\cref{fig:suppl_compare_editing_bagel}, and  \method can precisely understand and respond to the user's intentions, resulting in reliable editing outcomes.
We attribute this to the fact that our H-shape design with semantic tokens helps improve the semantic understanding capabilities of generative models.
%


\section{Additional Ablation Results}
\label{sec:More_Ablation}

\begin{figure}[t]
    \centering
    \includegraphics[width=0.99\linewidth]{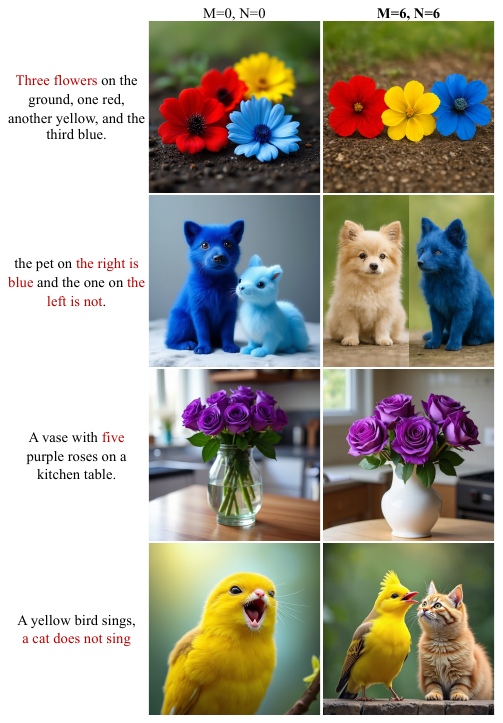}
    \vspace{-3mm}
    \caption{Qualitative ablation study on the number of skipped layers under
the \textbf{7B+4B} setting.}
    \label{fig:Supp_Ablation_MN_7B}
\end{figure}

\vspace{1mm}
\noindent
\textbf{Overfitting Phenomenon under 7B+4B Settings.}  To further verify that the fully layer-by-layer connected method may easily overfit the shallow features of the understanding expert, we conduct additional experiments under the 7B+4B configuration. From the result in~\cref{fig:Supp_Ablation_feature_analysis}, it can be seen that M=N=0 easily leads to overfitting of shallow features, while the setting of \method focuses primarily on the features of the intermediate semantic layers, resulting in better semantic coherence.
We also show the quantitative results under the 7B+4B setting in~\cref{tab:ablation_mn_7B_4B}, we can find that the mid-layer bridge (M=N=6) performs better than the baseline counterpart (M=N=0). 
In addition, the qualitative results in~\cref{fig:Supp_Ablation_MN_7B} demonstrate that M=N=0 may ignore some high-level semantics in textual prompts.
The conclusions are consistent with those under the 0.5B+4B setting in the main document.

\begin{table}[t]
    \centering
    \small
    \caption{Ablation study on the number of skipped layers under the \textbf{7B+4B} setting.  LLM re-writer is not used on GenEval.
    }
    \vspace{-3mm}
    \setlength\tabcolsep{17.0pt}
    \begin{tabular}{l|c|c}
    \shline
      \hspace{-5mm}Setting   &  DPG-Bench & GenEval \\
      \shline
       \hspace{-5mm}M=0, \hspace{1mm} N=0  & 83.21 & 0.80 \\
       \hspace{-5mm}M=6, \hspace{1mm} N=6 \textbf{(Ours)}  & \textbf{85.23} & \textbf{0.83} \\
       \shline
    \end{tabular}
    \label{tab:ablation_mn_7B_4B}
\end{table}

\begin{figure}[t]
    \centering
    \includegraphics[width=0.99\linewidth]{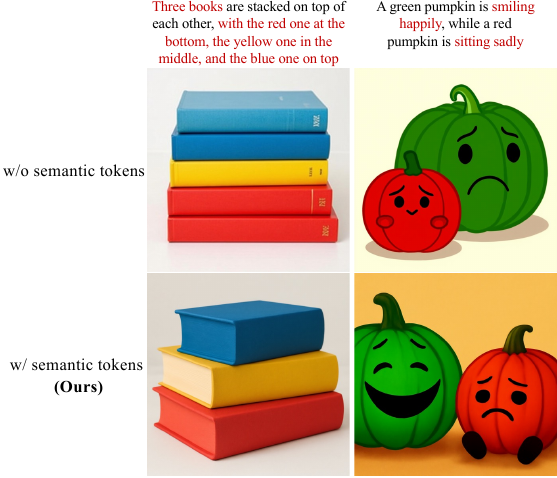}
    \vspace{-3mm}
    \caption{Qualitative ablation study on the effect of the proposed semantic reconstruction tokens. }
    \label{fig:Supp_Ablation_semantic_tokens}
\end{figure}

\vspace{1mm}
\noindent
\textbf{Effect of Semantic Reconstruction Tokens.} 
To qualitatively analyze the efficacy of the proposed semantic reconstruction tokens, we visualize some examples in~\cref{fig:Supp_Ablation_semantic_tokens}.
From the results, we can notice that incorporating semantic reconstruction tokens helps to enhance the ability to perceive position and attributes.
As shown in~\cref{tab:ablation_number_tokens}, we further conduct an ablation study on the number of learnable semantic tokens and find that 16 tokens achieve excellent performance.
\vspace{-1mm}

\begin{table}[t]
    \centering
    \small
    \caption{Ablation study on the number of learnable semantic tokens under the {0.5B+4B} setting.  
    }
    \vspace{-3mm}
    \setlength\tabcolsep{24.0pt}
    \begin{tabular}{l|c|c}
    \shline
      \hspace{-7mm}Setting   &  DPG-Bench & GenEval \\
      \shline
       \hspace{-7mm}4 tokens  & 79.82  & 0.65 \\
       \hspace{-7mm}16 tokens  & \textbf{80.03} & \textbf{0.66} \\
       \hspace{-7mm}36 tokens  & {79.83} & \textbf{0.66} \\
       \shline
    \end{tabular}
    \label{tab:ablation_number_tokens}
\end{table}


\end{document}